\documentclass[a4paper,12pt,reqno]{amsart} 
\usepackage{graphicx}
\graphicspath{ {images/} }
\usepackage[english]{babel}
\usepackage{amssymb}
\usepackage{tikz-cd}
\usepackage{tikz}
\usepackage{amsmath}
\usepackage{dsfont}
\usepackage{shuffle}
\usepackage{afterpage}
\usepackage{url}
\usepackage{setspace}
\usepackage{booktabs}
\usepackage{multirow}
\usepackage{placeins}
\usepackage[utf8]{inputenc}

\usepackage{mathrsfs}
\usepackage{mathbbol}
\usepackage[margin=1in]{geometry}
\usepackage{algorithm,algorithmicx,algpseudocode}

\usepackage[bookmarks=true,
bookmarksnumbered=true, breaklinks=true,
pdfstartview=FitH, hyperfigures=false,
plainpages=false, naturalnames=true,
colorlinks=true,pagebackref=true,
pdfpagelabels]{hyperref}

\usepackage[all]{xy}
\usetikzlibrary{knots} 
\usepackage[lite]{amsrefs}
\usepackage{bbm}

\title[NIO for inverse problems]{Neural Integral Operators for Inverse Problems: \\
An Operator-Learning Framework for Small-Sample Spectroscopic Classification}

\author[Zappala]{Emanuele Zappala$^*$}
\address{Department of Mathematics and Statistics, Idaho State University,
Physical Science Complex, 921 S. 8th Ave., Stop 8085, Pocatello, ID 83209, USA} 
\email{emanuelezappala@isu.edu, ORCiD 0000-0002-9684-9441}

\author[Giola]{Alice Giola}
\address{Department of Mathematics and Statistics, Idaho State University,
Physical Science Complex, 921 S. 8th Ave., Stop 8085, Pocatello, ID 83209, USA}
\email{alicegiola@isu.edu}

\author[Kramer]{Andreas Kramer}
\address{Department of Computer Science, Idaho State University,
921 S. 8th Ave Mail Stop 8060, Pocatello, ID 83209-8023, USA}
\email{andreaskramer@isu.edu}

\author[Acharya]{Saugat Acharya}
\address{Department of Computer Science, Idaho State University,
921 S. 8th Ave Mail Stop 8060, Pocatello, ID 83209-8023, USA}
\email{saugatacharya@isu.edu}

\author[Greco]{Enrico Greco$^*$}
\address{Institute for the Advanced Study of Culture and the Environment (IASCE), University of South Florida, 4202 E Fowler Ave, Tampa, FL 33620, USA}
\email{enrico.greco@live.it, ORCiD 0000-0003-1564-4661}

\thanks{$^*$Corresponding Author}

\begin{document}

\begin{abstract}
Learning maps between function spaces with a strong inductive bias is a central challenge in soft computing, especially when training data are scarce and standard deep architectures overfit. We introduce a \emph{neural integral operator} (NIO) framework based on integral equations of the first kind, in which the Urysohn kernel of the operator is parameterized by a feed-forward network~$G_{\theta_G}$ and the latent function is produced by a convolutional encoder~$E_{\phi_E}$, both trained jointly end-to-end via cross-entropy loss. The integral defining the learned operator is approximated by Monte Carlo sampling, which we argue acts as an implicit stochastic regularizer operating at the level of the integrand and complementing parameter-level regularizers such as weight decay and dropout. We benchmark the framework on three real-world spectroscopic classification tasks (FT-IR fruit purees, NIR meat, NIR textiles) of varying size and complexity, against traditional machine learning (decision tree, support vector machine, with and without UMAP) and modern deep learning baselines (FFNN, CNN+FFNN, shallow CNN, transformer). The proposed NIO is consistently among the top two performing models across all datasets and metrics, achieves the best results on the most challenging small-and-complex dataset (Textile), and yields lower performance variance than competing deep models in the small-data regime. The results suggest that operator-learning architectures with stochastic numerical integration are a viable soft-computing strategy for inverse problems in spectroscopy when conventional deep learning approaches are limited by data scarcity.
\end{abstract}


\maketitle

\section{Introduction}\label{sec:introduction}

A recurring difficulty in soft computing is the design of learning architectures that admit a strong inductive bias on the structure of the mapping to be learned, while remaining flexible enough to fit data drawn from complex generative processes. When the input to the system is naturally a function of a continuous variable, rather than a fixed-dimensional vector, the learning target itself is best modelled as an \emph{operator} between infinite-dimensional spaces. \emph{Neural operators} \cite{Lu2021,LiFNO,KovachkiNO} have recently emerged as a family of architectures designed for this setting, combining tools from functional analysis with deep learning. Within this family, the operator is parameterized by neural networks and learned from finite samples of input/output function pairs, with universal-approximation guarantees of varying strength depending on the architecture and functional spaces considered.

While the bulk of the neural operator literature has focused on forward problems governed by partial differential equations, comparatively little attention has been devoted to the use of operator learning for \emph{inverse problems}, where the unknown is the input function (or a property thereof) and the observation is its image under a forward operator. A natural mathematical formulation for such inverse problems is given by \emph{integral equations of the first kind} \cite{groetsch2007integral}, in which an unknown function~$u$ is related to an observed signal~$f$ via integration against a kernel. Equations of this type have been used for decades in inverse problems arising in geophysics, climatology, immunology and analytical chemistry, but their integration into the neural operator framework remains essentially unexplored. The most directly related prior work is the Neural Integral Equation (NIE) framework \cite{ANIE} and its integro-differential counterpart \cite{NIDE}, which address \emph{second-kind} equations for dynamical systems. To the best of our knowledge, no existing neural operator method explicitly targets first-kind integral equations as a vehicle for inverse classification or regression.

In this paper, we propose a Neural Integral Operator (NIO) for inverse problems formulated as integral equations of the first kind. The three main contributions of the work are the following:
\begin{itemize}\setlength\itemsep{0.25em}
\item[\textbf{(C1)}] We introduce an end-to-end trainable architecture in which a learned Urysohn kernel $G_{\theta_G}$ and a convolutional encoder $E_{\phi_E}$ jointly parameterize the unknown integral operator $T$, with classification labels recovered by evaluating $T u_{\phi_E}$ at a learned readout point $\sigma_E$ produced by the encoder.
\item[\textbf{(C2)}] We approximate the integral defining $T$ via Monte Carlo sampling and argue that the resulting stochastic numerical scheme acts as an \emph{implicit regularizer} on the operator, analogous in spirit to dropout and stochastic noise injection but operating at the level of the integrand. Empirically, the NIO is markedly more robust to overfitting than feed-forward, convolutional and transformer baselines of comparable parameter count, especially on small datasets.
\item[\textbf{(C3)}] We provide an extensive empirical study on three real-world spectroscopic classification datasets of varying size and complexity, against classical chemometric, machine learning and modern deep learning baselines, supported by tests over 10 Monte Carlo cross-validation splits.
\end{itemize}

\paragraph{\textbf{Application domain as testbed}}
We use spectroscopic classification as a testbed for the proposed framework for two reasons. First, spectroscopy is a paradigmatic instance of an inverse problem with a strong functional structure: spectra are naturally modelled as sampled functions over a continuous spectral domain, and the assignment of a class label corresponds to recovering structural or compositional information from an indirect observation \cite{mark2010comparison,houston2020robust}. Second, spectroscopic datasets are typically of modest size, providing a stress test for the small-data behaviour of operator-learning architectures. For comprehensive reviews of machine learning and deep learning approaches to NIR, FT-IR and Raman spectroscopy, the reader is referred to \cite{zhang2022review,MishraDLNIR,luo2022deep,ContrerasXAI2024}; quantitative interpretation of spectra has historically relied on chemometric calibration models such as PLS regression \cite{Wold2001}. Several strategies have been developed to alleviate the small-sample bottleneck, including transfer learning \cite{michelucci2024deep,kalatzis2023advanced,QuantNIRDL}, self-supervised pre-training \cite{ZhaoNIRSSL,AbdalmalakfNIRS}, and specialized 1D-CNN or hybrid CNN--Transformer architectures \cite{RamanNet,RSTransformer}. Our work is complementary to these efforts: rather than augment the data or transfer knowledge from a source domain, we introduce an inductive bias at the level of the architecture (the operator structure) and at the level of the numerical integration scheme. While we adopt spectroscopic language in the experimental sections, the proposed framework is domain-agnostic and applies wherever an inverse problem can be cast as an integral equation of the first-kind.

\begin{figure}[t]
    \centering
    \includegraphics[width=\linewidth]{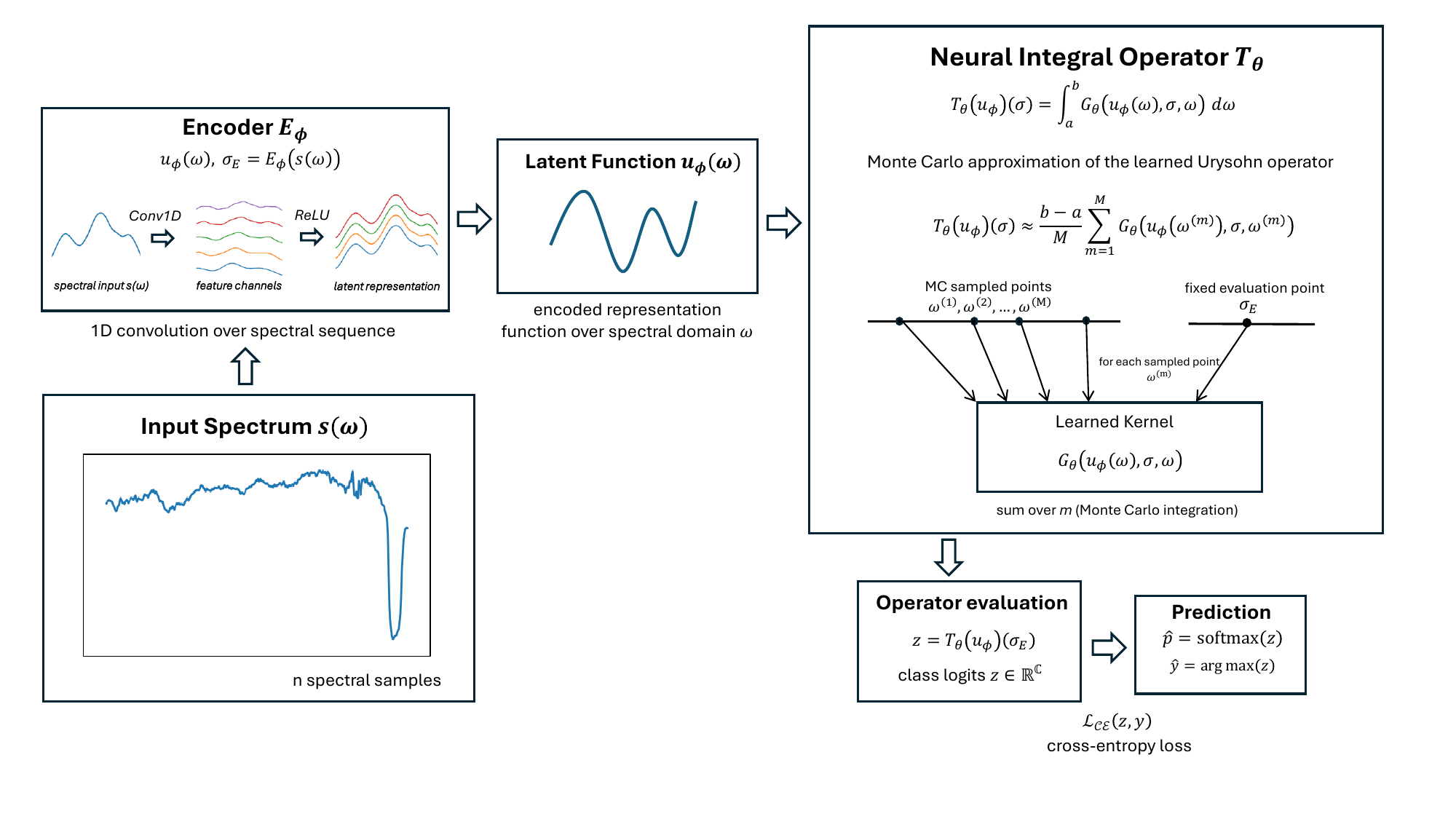}
    \caption{\textbf{Neural integral operator pipeline.}
    Given a one-dimensional spectrum $s(\omega)$, a convolutional encoder $E_{\phi_E}$ extracts local features and maps the input into a latent function $u_{\phi_E}(\omega)$, together with a learned evaluation coordinate $\sigma_E$. This representation is passed to a neural integral operator $T_{\theta_G}$, defined via a learned Urysohn kernel $G_{\theta_G}\bigl(u_{\phi_E}(\omega), \sigma, \omega\bigr)$, which captures nonlocal relationships by integrating over the spectral domain. In practice, the integral is approximated via Monte Carlo sampling over points $\{\omega^{(m)}\}_{m=1}^{M}$, whose contributions are aggregated at the evaluation point $\sigma_E$. The operator output produces class logits $z = (T_{\theta_G} u_{\phi_E})(\sigma_E)$, which are transformed into class probabilities via softmax. The model is trained end-to-end using cross-entropy loss. The symbols $\phi$ and $\theta$ appearing inside the figure correspond, respectively, to the encoder parameters $\phi_E$ and the kernel parameters $\theta_G$ used in the main text.}
    \label{fig:diagram_spectroscopy}
\end{figure}

A schematic overview of the full pipeline is shown in Figure~\ref{fig:diagram_spectroscopy}.

\paragraph{\textbf{Organization of the paper}}
The remainder of the paper is organized as follows. Section~\ref{sec:related} reviews related work on neural operators and inverse problems and positions our contribution with respect to the existing literature. Section~\ref{sec:methods} introduces the neural integral operator framework and presents the training algorithm, with explicit notation for the encoder and kernel parameters. Section~\ref{sec:experiments} describes the datasets, the baselines, and the preprocessing pipeline and the experimental protocol. Section~\ref{sec:results} reports the empirical results and the class-conditional diagnostics. Section~\ref{sec:limitations} discusses the limitations of the present study and outlines future work, including extensions to regression and to alternative quadrature schemes. Section~\ref{sec:conclusions} concludes. Code, and Jupyter notebooks are available at \url{https://github.com/emazap7/spectroscopy_integral_operator}.

\section{Related Work}\label{sec:related}

\subsection{Neural operators and operator learning}
Neural operators are a family of machine learning architectures designed to learn maps between function spaces from finite-sample data. DeepONet \cite{Lu2021} introduced a branch--trunk decomposition that approximates a nonlinear operator as an inner product between a representation of the input function and a representation of the evaluation point, supported by a universal approximation theorem for operators. Fourier Neural Operators \cite{LiFNO} subsequently parameterized the operator kernel in the spectral domain, achieving substantial efficiency gains for problems with translation invariance and smooth kernels. A unifying mathematical framework, including approximation, generalization and stability properties for several neural operator families, is provided by Kovachki et al.\ \cite{KovachkiNO}. Our previous work introduced Neural Integral Equations (NIE) \cite{ANIE} and Neural Integro-Differential Equations \cite{NIDE} as operator-learning architectures for second-kind equations arising in dynamical systems, where a fixed-point equation is learned and solved during training. The present paper extends this line of work to \emph{first-kind} integral equations, which are the canonical formulation for inverse problems and which lack the contractive structure that simplifies the analysis of second-kind equations. A key feature of our formulation is that it exposes a direct connection between Monte Carlo numerical integration and implicit regularization, an aspect that has not been emphasized in earlier neural operator literature.

\subsection{Inverse problems via integral equations of the first kind}
Inverse problems, formulated as integral equations of type $T u = f$, where $T$ is a linear or nonlinear integral operator between Banach spaces, have a long history \cite{groetsch2007integral}, with classical examples ranging from Abel's gravity problem to system identification, communication engineering, paleoclimatology, immunology and polymer segmentation. The first-kind integral equation
\begin{equation*}
\int_\Omega G\bigl(u(\omega), \sigma, \omega\bigr)\, d\omega = f(\sigma)
\end{equation*}
with Urysohn kernel $G$ provides a flexible nonlinear template that encompasses linear and Hammerstein kernels as special cases. Classical numerical schemes for these equations are often ill-posed and can be tackled with Tikhonov regularization, truncated singular value decomposition, or other forms of bias--variance control. Our approach is methodologically different: rather than regularize a fixed, known operator, we \emph{learn} the operator (via its kernel) and the corresponding unknown function jointly from supervised data, transferring the burden of regularization to architectural choices (the operator structure) and to the stochastic numerical integration scheme.

\subsection{Deep learning for spectroscopy}
Reviews on machine learning and deep learning for NIR, FT-IR and Raman spectroscopy \cite{zhang2022review,MishraDLNIR,luo2022deep,ContrerasXAI2024} document the rapid adoption of convolutional, recurrent and, more recently, attention-based architectures. Specialized 1D-CNN architectures such as RamanNet \cite{RamanNet} and hybrid CNN--Transformer designs \cite{RSTransformer} have set strong baselines on Raman classification benchmarks, while feed-forward networks and CNN+FFNN combinations remain the most widely used backbones for routine FT-IR and NIR classification. Explainability of deep spectroscopic models is an active area of investigation \cite{ContrerasXAI2024}. In our experiments we include a feed-forward baseline, a CNN+FFNN baseline, a shallow CNN, and a transformer, which together provide a representative cross-section of the architectures most commonly considered in the spectroscopic deep learning literature within a comparable parameter budget. We discuss the implications of not including larger state-of-the-art models in Section~\ref{sec:limitations}.

\subsection{Learning from small spectroscopic datasets}
Small training sets remain a defining constraint for spectroscopic deep learning. Strategies developed in response include: (i) careful regularization of shallow or residual CNNs, often combined with chemometric preprocessing pipelines; (ii) transfer learning and domain adaptation, where models pre-trained on larger spectral corpora are fine-tuned on the target task \cite{michelucci2024deep,kalatzis2023advanced,QuantNIRDL}; and (iii) self-supervised pre-training, which has recently been shown to substantially improve small-sample NIR and fNIRS classification \cite{ZhaoNIRSSL,AbdalmalakfNIRS}. Our work is orthogonal and complementary to these strategies: the inductive bias introduced by the operator structure and the stochastic integration scheme reduces overfitting \emph{without} requiring auxiliary data, additional pretraining or task-specific preprocessing pipelines. In principle, the proposed NIO could be combined with self-supervised or transfer-learning strategies, and we identify this combination as a promising direction in Section~\ref{sec:limitations}.

\section{Neural Integral Operator: Methodology}\label{sec:methods}

This section presents the proposed Neural Integral Operator (NIO) framework in detail. We first formalize the inverse classification problem as a first-kind integral equation (Section~\ref{subsec:problem}), then describe the neural parameterization of the unknown function and the unknown kernel (Section~\ref{subsec:param}), the Monte Carlo numerical scheme used to evaluate the operator (Section~\ref{subsec:MC}), and the resulting training objective and algorithm (Section~\ref{subsec:training}). Implementation details, including the encoder and kernel architectures and the training hyperparameters, are summarized in Section~\ref{subsec:impl}. The final Section~\ref{subsec:MC_reg} analyses Monte Carlo integration as an implicit stochastic regularizer, which is a fundamental methodological observation of the paper.

\subsection{Problem formulation}\label{subsec:problem}

We consider supervised classification problems in which each input is a function $s: [a,b] \to \mathbb{R}$, sampled at a finite set of points $\omega_1, \ldots, \omega_n \in [a,b]$, and the corresponding label $\ell$ belongs to a finite set $\mathcal{Y} = \{1, \ldots, C\}$ of $C$ classes. In the experiments of Section~\ref{sec:experiments}, $s$ is a measured spectrum and $[a,b]$ is the wavelength or wavenumber interval over which the spectrum is recorded; $n$ is the number of spectral channels and depends on the instrument.

Following the inverse-problem template introduced in Section~\ref{sec:related}, we model the classifier as the evaluation of a learned integral operator at a learned readout point. Concretely, we postulate the existence of:
\begin{itemize}\setlength\itemsep{0.2em}
\item an unknown latent function $u: [a',b'] \to \mathbb{R}^d$, lying in a reduced spectral domain $[a',b'] \subseteq [a,b]$ of dimension $d \geq 1$;
\item an unknown nonlinear integral operator $T$ with Urysohn kernel $G$, such that
\begin{equation}\label{eqn:T_def}
(T u)(\sigma) \;=\; \int_{a'}^{b'} G\bigl(u(\omega), \sigma, \omega\bigr)\, d\omega \,\in\, \mathbb{R}^{C} ;
\end{equation}
\item a single readout point $\sigma_E \in [a',b']$ at which the operator is evaluated to produce the class logits,
\begin{equation}\label{eqn:logits}
z \;=\; (T u)(\sigma_E) \,\in\, \mathbb{R}^{C} .
\end{equation}
\end{itemize}
Class probabilities and predicted labels are recovered via softmax,
\begin{equation*}
\hat p \;=\; \mathrm{softmax}(z), \qquad \hat\ell \;=\; \arg\max_{c \in \{1,\ldots,C\}} z_c .
\end{equation*}
The function $u$, the kernel $G$, the dimension $d$, the reduced domain $[a',b']$ and the readout point $\sigma_E$ are all to be learned from data. The output of the operator is therefore a $C$-dimensional vector of logits, so that the abstract data function $f(\sigma) = (Tu)(\sigma)$ of Section~\ref{sec:related} is in fact a function $[a',b'] \to \mathbb{R}^{C}$, of which we retain only the value at $\sigma = \sigma_E$ for label prediction. This makes explicit how the abstract first-kind integral equation $Tu = f$ is instantiated in the concrete classifier: the right-hand side $f$ encodes the one-hot label information, and $\sigma_E$ is the single point at which the label-readout is enforced.

\subsection{Neural parameterization}\label{subsec:param}

We parameterize the unknown latent function $u$ and the unknown kernel $G$ by two neural networks with \emph{distinct} parameter sets, denoted $\phi_E$ and $\theta_G$ respectively, to avoid the notational overloading of the previous formulation.

\paragraph{\textbf{Encoder}}
A convolutional encoder
\begin{equation*}
E_{\phi_E}: \mathbb{R}^{n} \,\longrightarrow\, \mathbb{R}^{d \times n'} \times [a',b']
\end{equation*}
maps the input spectrum $s$, viewed as a vector of $n$ samples, to two outputs: a discretized latent function $\mathbf{u}_{\phi_E} \in \mathbb{R}^{d \times n'}$ over $n'$ points of the reduced spectral domain $[a',b']$, and a scalar readout coordinate $\sigma_E \in [a',b']$. The continuous latent function $u_{\phi_E}: [a',b'] \to \mathbb{R}^{d}$ is obtained from the discrete tensor $\mathbf{u}_{\phi_E}$ by piecewise linear interpolation between sample points, which makes $u_{\phi_E}(\omega)$ differentiable almost everywhere with respect to $\phi_E$. Architectural details (number of layers, channels, kernel sizes) are reported in Sec.~\ref{subsec:impl}; the role of $E_{\phi_E}$ is to combine local feature extraction (through 1D convolutions) with a learned reduction of the spectral domain, which significantly lowers the cost of the subsequent integration.

\paragraph{\textbf{Kernel}}
A feed-forward neural network
\begin{equation*}
G_{\theta_G}: \mathbb{R}^{d} \times \mathbb{R} \times \mathbb{R} \,\longrightarrow\, \mathbb{R}^{C}
\end{equation*}
parameterizes the Urysohn kernel. Its inputs are the latent function value $u_{\phi_E}(\omega) \in \mathbb{R}^{d}$, the evaluation variable $\sigma \in [a',b']$, and the integration variable $\omega \in [a',b']$; its output is a $C$-dimensional contribution to the integrated logits. The integration variable $\omega$ is provided to the kernel as an explicit input, so that $G_{\theta_G}$ can encode spatial structure along the spectral axis (in particular, asymmetric or non-translation-invariant interactions between $\sigma$ and $\omega$) that a purely convolutional model would have to capture indirectly.

The learned integral operator is then
\begin{equation}\label{eqn:T_theta_def}
\bigl(T_{\theta_G} u_{\phi_E}\bigr)(\sigma) \;=\; \int_{a'}^{b'} G_{\theta_G}\bigl(u_{\phi_E}(\omega), \sigma, \omega\bigr)\, d\omega ,
\end{equation}
and the classification logits for a single input spectrum $s$ are recovered as $z = (T_{\theta_G} u_{\phi_E})(\sigma_E)$. The parameter pair $(\theta_G, \phi_E)$ is trained jointly end-to-end with a cross-entropy loss; no prior knowledge of the functional dependence of the labels on the spectra is required, since the structure of the inverse problem is entirely encoded in the architecture and the inverse mapping is learned from data.

\subsection{Monte Carlo numerical integration}\label{subsec:MC}

The integral~\eqref{eqn:T_theta_def} cannot be evaluated in closed form, since both $u_{\phi_E}$ and $G_{\theta_G}$ are parameterized by neural networks. We approximate it via plain (uniform) Monte Carlo integration:
\begin{equation}\label{eqn:MC}
\bigl(T_{\theta_G} u_{\phi_E}\bigr)(\sigma) \;\approx\; \frac{b'-a'}{M} \sum_{m=1}^{M} G_{\theta_G}\bigl(u_{\phi_E}(\omega^{(m)}), \sigma, \omega^{(m)}\bigr) ,
\end{equation}
where $\omega^{(1)}, \ldots, \omega^{(M)}$ are drawn independently and uniformly from $[a',b']$. The estimator is unbiased and has variance proportional to $1/M$. We use $M_{\mathrm{train}} = 2000$ samples during training and $M_{\mathrm{test}} = 5000$ samples at evaluation. Numerical integration is implemented through the \texttt{torchquad} library~\cite{gomez2021torchquad}, which provides differentiable PyTorch Monte Carlo quadrature with gradient flow through the integrand: the sample locations $\omega^{(m)}$ themselves do not require gradients, but each evaluation $G_{\theta_G}(\cdot, \sigma, \omega^{(m)})$ contributes to $\nabla_{(\theta_G, \phi_E)} \mathcal{L}$ through the chain rule (i.e. backpropagation).

The asymmetry $M_{\mathrm{train}} < M_{\mathrm{test}}$ is intentional. During training, a moderate $M_{\mathrm{train}}$ keeps the per-batch compute and memory cost tractable while still providing an unbiased estimator of the operator; in addition, as discussed in Sec.~\ref{subsec:MC_reg}, the residual stochasticity in~\eqref{eqn:MC} introduces a beneficial perturbation in the loss surface. During evaluation, where gradients are not required and the cost of additional samples is amortized over a much smaller number of forward passes, the variance of the estimator can be reduced cheaply by increasing $M$, yielding a more stable predictor and lower variance of the test metrics, as we have consistently observed in our numerical experiments.

\subsection{Training objective and algorithm}\label{subsec:training}

Training data consist of $N$ pairs $\{(s_i, \ell_i)\}_{i=1}^{N}$, where $s_i$ is a sampled spectrum and $\ell_i \in \{1, \ldots, C\}$ is the corresponding class label, encoded as a one-hot vector $y_i \in \{0,1\}^{C}$. For each example the model computes, in order: (i) the encoded latent function and readout point $(\mathbf{u}_{\phi_E}^{(i)}, \sigma_E^{(i)}) = E_{\phi_E}(s_i)$; (ii) a fresh set of Monte Carlo nodes $\{\omega^{(m,i)}\}_{m=1}^{M_{\mathrm{train}}}$ drawn uniformly from $[a',b']$; (iii) the Monte Carlo logits via~\eqref{eqn:MC},
\begin{equation*}
z_i \;=\; \frac{b'-a'}{M_{\mathrm{train}}} \sum_{m=1}^{M_{\mathrm{train}}} G_{\theta_G}\bigl(u_{\phi_E}^{(i)}(\omega^{(m,i)}),\, \sigma_E^{(i)},\, \omega^{(m,i)}\bigr), 
\end{equation*}
where the latent function is interpolated to allow arbitrary evaluations during Monte Carlo integration. 
The training loss is the standard cross-entropy on the resulting logits,
\begin{equation}\label{eqn:CE_loss}
\mathcal{L}_{\mathrm{CE}}(\theta_G, \phi_E) \;=\; -\frac{1}{N}\sum_{i=1}^{N} \sum_{c=1}^{C} y_{i,c}\,\log\bigl(\mathrm{softmax}(z_i)_c\bigr) ,
\end{equation}
optimized jointly with respect to $(\theta_G, \phi_E)$ by Adam stochastic gradient descent. The full training procedure is given in Algorithm~\ref{algo:NIEIP}.

\begin{algorithm}[h!]
    \caption{Training the Neural Integral Operator (NIO).}
    \label{algo:NIEIP}
    \begin{algorithmic}[1]
        \Require Training set $\{(s_i, y_i)\}_{i=1}^{N}$; Monte Carlo budget $M_{\mathrm{train}}$; reduced domain $[a',b']$; learning rate $\eta$; number of epochs $E$; minibatch size $B$.
        \Ensure Trained parameters $(\theta_G, \phi_E)$.
        \State Initialize $\theta_G, \phi_E$ randomly.
        \For{$\mathrm{epoch} = 1, \ldots, E$}
            \For{each minibatch $\mathcal{B} \subset \{1, \ldots, N\}$ of size $B$}
                \For{each $i \in \mathcal{B}$}
                    \State $(\mathbf{u}_{\phi_E}^{(i)}, \sigma_E^{(i)}) \gets E_{\phi_E}(s_i)$ \Comment{encode spectrum}
                    \State Draw $\omega^{(1,i)}, \ldots, \omega^{(M_{\mathrm{train}},i)} \stackrel{\text{i.i.d.}}{\sim} \mathcal{U}([a',b'])$
                    \State $z_i \gets \dfrac{b'-a'}{M_{\mathrm{train}}} \displaystyle\sum_{m=1}^{M_{\mathrm{train}}} G_{\theta_G}\bigl(u_{\phi_E}^{(i)}(\omega^{(m,i)}), \sigma_E^{(i)}, \omega^{(m,i)}\bigr)$ \Comment{MC logits}
                \EndFor
                \State $\mathcal{L} \gets -\dfrac{1}{|\mathcal{B}|} \displaystyle\sum_{i \in \mathcal{B}} \sum_{c=1}^{C} y_{i,c} \log \mathrm{softmax}(z_i)_c$ \Comment{cross-entropy}
                \State $(\theta_G, \phi_E) \gets (\theta_G, \phi_E) - \eta\, \nabla_{(\theta_G,\phi_E)}\, \mathcal{L}$ \Comment{Adam update}
            \EndFor
            \State Optionally apply early stopping on a held-out validation split.
        \EndFor
    \end{algorithmic}
\end{algorithm}

\paragraph{\textbf{Connection to the inverse problem}}
Equation~\eqref{eqn:CE_loss} is formally a supervised classification loss, but it admits a direct interpretation as a regularized approximate solution of the abstract first-kind integral equation $T u = f$ of Section~\ref{sec:related}. The unknown data function $f$ is identified with the (smoothed) one-hot label function $y_i$ at the single point $\sigma_E^{(i)}$, the unknown function $u$ is parameterized via $u_{\phi_E}$, and the unknown operator $T$ is parameterized via $G_{\theta_G}$; the cross-entropy provides a probabilistically motivated discrepancy on the predicted logits, while regularization is provided implicitly by the architecture (operator form) and by the stochasticity of the Monte Carlo estimator (Section~\ref{subsec:MC_reg}). This perspective makes explicit that the proposed framework is not restricted to classification: replacing the cross-entropy by mean squared error and the softmax by the identity yields a regression formulation in which $f$ is a continuous target signal, an extension we discuss in Section~\ref{sec:limitations}.

\subsection{Implementation details}\label{subsec:impl}

The full pipeline is implemented in PyTorch, with numerical integration handled by \texttt{torchquad}~\cite{gomez2021torchquad}. The architectures of the encoder $E_{\phi_E}$ and the kernel $G_{\theta_G}$, together with all training hyperparameters, are summarized in Table~\ref{tab:hyperparams}. The same hyperparameters are used across all three datasets, with the sole exception of the input length $n$, which is dictated by the spectral length of each dataset (see Table~\ref{tab:dataset_overview}).

\begin{table}[h!]
\centering
\caption{Architecture and training hyperparameters for the proposed Neural Integral Operator. The same configuration is used across all three datasets, with the input length $n$ adapted to the spectral resolution of each dataset.}
\label{tab:hyperparams}
\begin{tabular}{lll}
\toprule
Component & Parameter & Value \\
\midrule
\multirow{5}{*}{Encoder $E_{\phi_E}$} 
    & Type & 1D convolutional stack + linear head \\
    & Number of Conv1D layers & 3 \\
    & Channels (in $\to$ out) & $1 \to 16 \to 32 \to 64$ \\
    & Kernel sizes & $5, 3, 3$ \\
    & Activation & ReLU \\
\midrule
\multirow{3}{*}{Latent representation}
    & Latent dimension $d$ & 8 \\
    & Reduced domain points $n'$ & 64 \\
    & Reduced domain $[a',b']$ & $[0,1]$ (normalized) \\
\midrule
\multirow{4}{*}{Kernel $G_{\theta_G}$}
    & Type & Multi-layer perceptron \\
    & Hidden layers $\times$ width & $4 \times 128$ \\
    & Activation & $\tanh$ \\
    & Output dimension & $C$ (number of classes) \\
\midrule
\multirow{6}{*}{Training}
    & Optimizer & Adam \\
    & Learning rate $\eta$ & $10^{-3}$ \\
    & Weight decay & $10^{-4}$ \\
    & Batch size $B$ & 32 \\
    & Maximum epochs $E$ & 300 \\
    & Early stopping patience & 30 epochs (on validation loss) \\
\midrule
\multirow{2}{*}{Monte Carlo}
    & $M_{\mathrm{train}}$ (training samples) & 2000 \\
    & $M_{\mathrm{test}}$ (evaluation samples) & 5000 \\
\bottomrule
\end{tabular}
\end{table}

The total number of trainable parameters is dominated by the kernel MLP and is in the order of $\sim 10^{5}$, comparable to the FFNN and CNN+FFNN baselines used for comparison (see Sec.~\ref{sec:experiments}). All experiments are run on a single GPU; one training run on the Fruit dataset takes approximately 25 minutes, and proportionally less on the smaller Meat and Textile datasets. We emphasize that the elevated $M_{\mathrm{test}}$ does not significantly impact wall-clock evaluation time, since no gradients need to be propagated; this is the main practical motivation for the asymmetric training/evaluation budget of Section~\ref{subsec:MC}.

\subsection{Monte Carlo integration as implicit stochastic regularization}\label{subsec:MC_reg}

A central element of the proposed framework, and arguably its most distinctive feature for the soft-computing reader, is that the Monte Carlo approximation~\eqref{eqn:MC} is \emph{not merely a numerical device} but plays the role of an implicit stochastic regularizer, acting at the level of the integrand. We give here an informal argument for this claim; a formal analysis, including a precise comparison with the equivalent kernel viewpoint of dropout, is deferred to future work.

\paragraph{\textbf{Variance perturbation of the loss}}
For a fixed minibatch and a fixed parameter pair $(\theta_G, \phi_E)$, the Monte Carlo logits used in~\eqref{eqn:CE_loss} are random variables, since they depend on the freshly sampled nodes $\{\omega^{(m,i)}\}$. By the central limit theorem, their distribution around the true integral~\eqref{eqn:T_theta_def} is approximately Gaussian with covariance proportional to $1/M_{\mathrm{train}}$. Consequently, the gradient $\nabla_{(\theta_G,\phi_E)} \mathcal{L}_{\mathrm{CE}}$ used by Adam at each step is a noisy unbiased estimator of the gradient of the population loss. Stochastic gradient noise of this kind is well known to bias optimization towards flatter minima of the loss surface and to act as a regularizer on the learned function~\cite{SrivastavaDropout}.

\paragraph{\textbf{Analogy with dropout, weight noise and data augmentation}}
Dropout regularizes a network by injecting Bernoulli noise into hidden activations during training; data augmentation injects noise (or structured perturbations) into the inputs; weight noise injects Gaussian noise into the parameters. The Monte Carlo estimator in~\eqref{eqn:MC} injects what we call \emph{integrand noise}: at each forward pass, the operator $T_{\theta_G}$ is evaluated on a fresh random sub-sample of $[a',b']$ rather than on a fixed grid. The resulting variance in the logits is parameter- and input-dependent, but its overall magnitude can be tuned through $M_{\mathrm{train}}$: smaller $M_{\mathrm{train}}$ yields stronger stochastic regularization at the cost of higher gradient variance.

\paragraph{\textbf{Why deterministic quadrature is not equivalent}}
A natural alternative is to replace Monte Carlo with a deterministic quadrature rule (e.g.\ trapezoidal or Gauss--Legendre). Such a rule eliminates the stochastic noise in the integrand, restoring a deterministic loss landscape on each example. While this is useful at inference time, during training it removes the stochastic regularizer described above. In preliminary experiments not reported in the main text we observed that deterministic quadrature with comparable computational cost yields faster initial convergence but worse generalization in the small-data regime, consistent with the regularization interpretation. The choice of a relatively large $M_{\mathrm{train}} = 2000$ in our experiments ensures that the bias of the estimator is negligible, while the residual stochasticity continues to contribute to regularization; at evaluation time we increase $M_{\mathrm{test}}$ to $5000$ specifically to reduce variance in the predictions, where regularization is no longer the goal.

\paragraph{\textbf{Empirical consequences}}
The practical consequence of this analysis is that the proposed NIO is expected to exhibit two distinctive behaviours, both of which are observed in the experiments of Section~\ref{sec:results}: (i) lower performance variance across cross-validation splits compared with deep baselines of comparable parameter count, and (ii) reduced overfitting in the small-data regime, where dropout and weight decay applied to standard CNN, FFNN or transformer baselines do not fully prevent the degradation in test accuracy.

\section{Experimental Setup}\label{sec:experiments}

In this section we describe the datasets used to validate the proposed Neural Integral Operator framework (Section~\ref{subsec:datasets}), the preprocessing pipeline (Section~\ref{subsec:prepro}), the baseline classifiers and their hyperparameters (Section~\ref{subsec:baselines}), and the cross-validation protocol (Section~\ref{subsec:protocol}).

\subsection{Datasets}\label{subsec:datasets}

\begin{figure}[t]
    \centering
    \includegraphics[width=\linewidth]{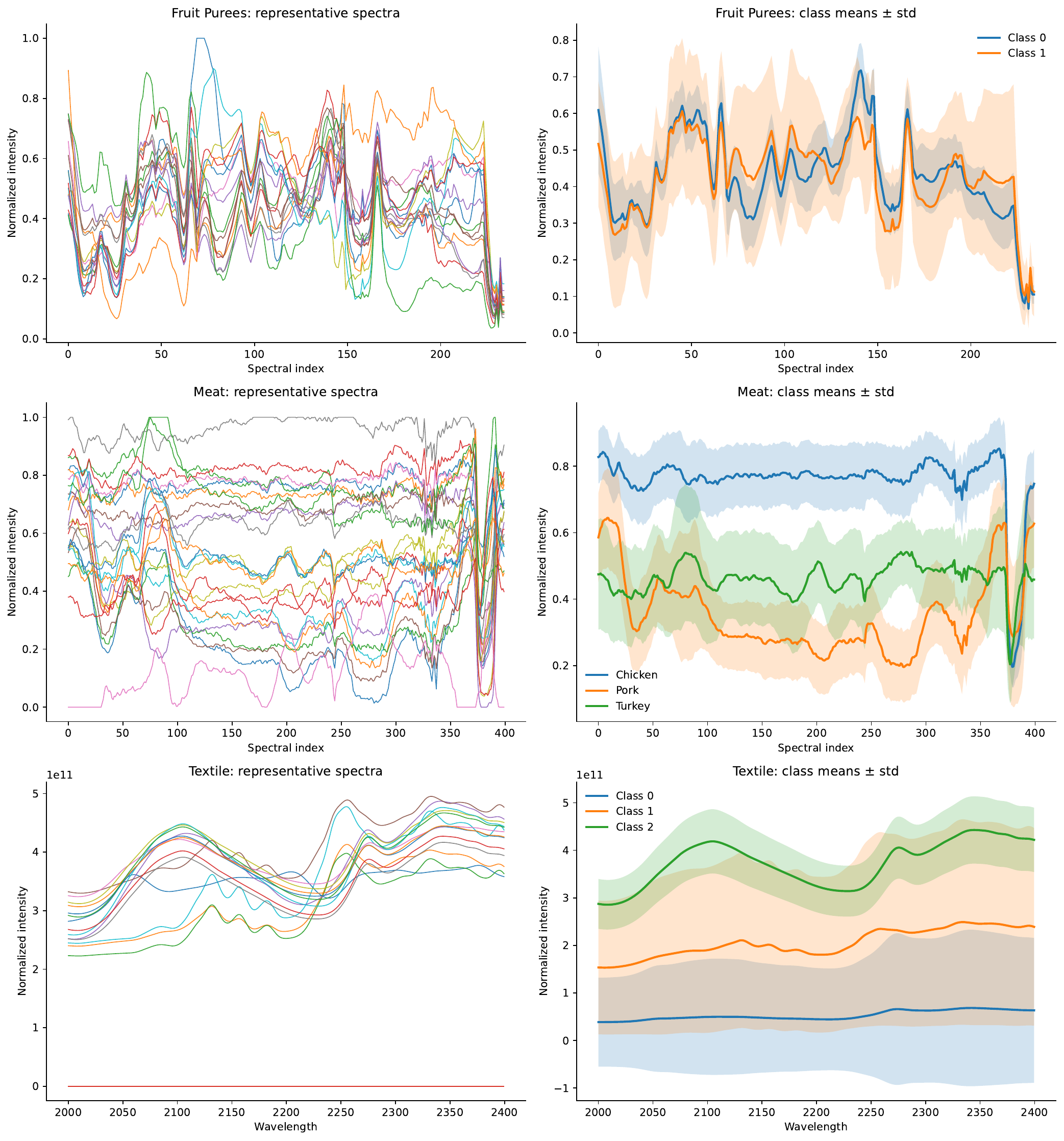}
    \caption{\textbf{Spectral distributions and class-conditional statistics across the three benchmark datasets.}
    Left column: a random subset of spectra per class, visualizing intra-class variability. Right column: class-conditional mean spectra with shaded $\pm 1$ standard deviation bands. Top: Fruit Purees (binary). Middle: Meat (Chicken, Pork, Turkey). Bottom: Textile (three classes). For Fruit and Meat, spectra are min--max normalized per spectral channel across the dataset, after truncation of Meat to the first 400 indices. For Textile, spectra are cropped to the wavelength window 2000--2400 nm and normalized per spectrum with an $\ell_{p}$ norm ($p = 1.53$). The class-conditional means and overlaps shown here are useful diagnostics for the difficulty pattern observed in the benchmark of Sec.~\ref{sec:results}.}
    \label{fig:representative_spectra}
\end{figure}

We evaluate the proposed NIO and the baselines on three real-world spectroscopic classification datasets that span different spectral techniques, sample types, dataset sizes and class balances. Summary statistics are given in Table~\ref{tab:dataset_overview}; representative spectra and class-conditional means are shown in Figure~\ref{fig:representative_spectra}.

\begin{table}[h!]
\centering
\caption{Dataset overview. ``Spectral length'' refers to the number of discretized channels after preprocessing (Section~\ref{subsec:prepro}). Train, Val, Test report the absolute sample counts in one representative split; class counts refer to the full dataset before splitting.}
\label{tab:dataset_overview}

\resizebox{\textwidth}{!}{
\begin{tabular}{lrrrrrrrr}
\toprule
Dataset & Samples & Classes & Spectral Length & Class Counts & Train & Val & Test \\
\midrule
Fruit Purees & 983 & 2 & 235 & 351 / 632     & 791 & 96 & 96 \\
Meat         & 120 & 3 & 400 & 40 / 40 / 40  & 80  & 20 & 20 \\
Textile      & 223 & 3 & 800 & 54 / 123 / 46 & 179 & 22 & 22 \\
\bottomrule
\end{tabular}
}
\end{table}

\paragraph{\textbf{Fruit Purees (FT-IR)} \cite{holland1998use}}
The Fruit Purees dataset~\cite{holland1998use} consists of $N = 983$ FT-IR spectra of strawberry and non-strawberry fruit purees, recorded on a Spectra-Tech MonitIR FT-IR spectrometer. Each spectrum is represented as 235 discretized intensity values over the corresponding wavenumber range. The classification task is binary: detection of adulteration of strawberry purees. The class distribution is imbalanced ($351$ strawberry vs.\ $632$ non-strawberry), as reflected in the class-conditional means of Figure~\ref{fig:representative_spectra} (top row). With $\sim 10^{3}$ examples, Fruit Purees is on the upper end of what is typically considered \emph{small} for deep learning in spectroscopy, and we use it as a comfort-zone benchmark where most architectures are expected to perform well.

\paragraph{\textbf{Meat (NIR)} \cite{naes1989leverage}}
The Meat dataset~\cite{naes1989leverage} contains $N = 120$ NIR spectra of three meat types (chicken, pork, turkey), with 40 samples per class. Each spectrum has 400 spectral channels. A subset of the samples has been intentionally adulterated by introducing starch and soy proteins, to simulate the kind of spectral outliers commonly encountered in adulterated meat samples. The class-conditional means in Figure~\ref{fig:representative_spectra} (middle row) show a clear separation between chicken (high intensity, flat profile) and the other two classes, while pork and turkey are visually less distinguishable, particularly in the central spectral region. The combination of small sample size and intentional outliers makes this a paradigmatic stress test for the robustness of small-data classifiers.

\paragraph{\textbf{Textile (NIR)}}
The Textile dataset contains $N = 223$ NIR spectra of three textile classes, with class counts $54 / 123 / 46$ (markedly imbalanced). The raw spectra cover wavelengths from 1100 to 2500 nm with 2800 discretized points; following preliminary inspection, we crop to the range 2000--2400 nm, retaining 800 spectral channels, since this window contains the most discriminative features. The intra-class variability of the textile spectra is substantial (Figure~\ref{fig:representative_spectra}, bottom row), and the class-conditional standard-deviation bands overlap significantly across the full retained range. The combination of small size, heavy class imbalance and high cross-class overlap makes this the most challenging of the three datasets, and the regime in which the regularization properties of the NIO are expected to be most beneficial.

\paragraph{\textbf{Retention of outliers}}
We emphasize that the spectral outliers introduced into the Meat dataset by adulteration are \emph{retained} in all training and test splits. Removing them would require either prior knowledge of which samples are adulterated or a separate outlier-detection step (e.g.\ robust covariance estimation), neither of which is generally available in deployment scenarios. The results reported in Section~\ref{sec:results} therefore reflect the behaviour of each model under realistic contamination conditions.

\subsection{Preprocessing}\label{subsec:prepro}

We adopt a deliberately minimal preprocessing pipeline, designed to avoid injecting dataset-specific prior knowledge:
\begin{itemize}\setlength\itemsep{0.2em}
\item \textbf{Fruit Purees and Meat:} per-feature (i.e.\ per spectral channel) min--max normalization across the dataset to the interval $[0,1]$, applied after truncation of the Meat spectra to the first 400 indices.
\item \textbf{Textile:} cropping to the wavelength window 2000--2400 nm, followed by per-spectrum $\ell_{p}$ normalization with $p = 1.53$, chosen as an intermediate value between the absolute-value-preserving $\ell_{1}$ and the energy-preserving $\ell_{2}$. We found this normalization to stabilize training across all baselines and adopt it uniformly on this dataset.
\end{itemize}
No baseline correction, SNV (Standard Normal Variate), MSC (Multiplicative Scatter Correction) or derivative-based transformations are applied, contrary to common chemometric practice. This choice ensures that the comparison between models is conducted on essentially raw spectra modulo normalization, and isolates the contribution of the learning algorithm itself; integration with more sophisticated chemometric pipelines is discussed in Section~\ref{sec:limitations}. The same preprocessing is applied identically across all models in the comparison.

\subsection{Baseline classifiers}\label{subsec:baselines}

We compare the proposed NIO against a representative set of baselines spanning traditional chemometric and machine-learning methods on the one hand, and modern deep architectures on the other hand.

\paragraph{Classical methods.}
\begin{itemize}\setlength\itemsep{0.2em}
\item \textbf{Decision Tree (DT).} A standard CART decision tree (\texttt{scikit-learn}), with depth and minimum-leaf-size hyperparameters optimized per dataset by a genetic algorithm.
\item \textbf{Support Vector Machine (SVM).} A multi-class SVM with RBF kernel; the penalty $C$ and the kernel parameter $\gamma$ are optimized per dataset by a genetic algorithm.
\item \textbf{DT + UMAP and SVM + UMAP.} The same DT and SVM classifiers applied on top of UMAP~\cite{mcinnes2018umap,becht2019dimensionality} embeddings of the spectra. UMAP is configured with $n_{\mathrm{components}} = 8$, $n_{\mathrm{neighbors}} = 15$, $\mathrm{min\_dist} = 0.1$, kept fixed across datasets. The motivation for including these baselines is to verify whether nonlinear dimensionality reduction prior to classical classifiers can compete with deep architectures in the small-data regime, as has been reported in related contexts~\cite{blanco2022strategies,del2023comparative}.
\end{itemize}

\paragraph{Deep-learning baselines.}
\begin{itemize}\setlength\itemsep{0.2em}
\item \textbf{FFNN.} A feed-forward neural network with 3 hidden layers of width 256, ReLU activations and dropout 0.3.
\item \textbf{CNN + FFNN.} A 1D-CNN feature extractor (3 Conv1D layers with channels $16 \to 32 \to 64$, kernel sizes $5, 3, 3$, ReLU) followed by a 2-layer MLP classifier of width 128.
\item \textbf{Shallow CNN.} An explicitly low-capacity 1D-CNN (2 Conv1D layers with channels $8 \to 16$, kernel sizes $5, 3$, ReLU), followed by global average pooling and a linear classifier. Included to probe the effect of model capacity on overfitting in the small-data regime.
\item \textbf{Tiny Transformer.} A transformer encoder with $L = 2$ layers, hidden dimension 64, 4 attention heads and dropout 0.1, applied to a learned token embedding of the spectrum (patch size 8). The output is mean-pooled and passed to a linear classifier.
\end{itemize}
All deep-learning baselines are trained with the same Adam configuration and the same early-stopping rule as the NIO (Table~\ref{tab:hyperparams}). Trainable parameter counts are matched within an order of magnitude ($10^{4}$--$10^{5}$), so that observed performance differences are attributable primarily to architectural inductive biases rather than to raw capacity.

\subsection{Evaluation protocol}\label{subsec:protocol}

We adopt Monte Carlo cross-validation with $K = 10$ random splits. Each split assigns 90\% of the samples to the training set (with a further 90/10 internal split into actual training and validation) and the remaining 10\% to the test set. The 10 splits are generated by random permutation of the dataset indices without stratification by class; this is a deliberately demanding protocol, since it allows class imbalance to vary across splits.

Performance is evaluated using accuracy and macro-averaged precision, recall and $F_{1}$ score; means and standard deviations are reported over the $K = 10$ splits. We report macro averages rather than micro averages because the latter are dominated by the majority class in imbalanced settings (Fruit Purees and Textile).

\section{Results and Discussion}\label{sec:results}

\subsection{Overall benchmark}

\begin{figure}[t]
    \centering
    \includegraphics[width=\linewidth]{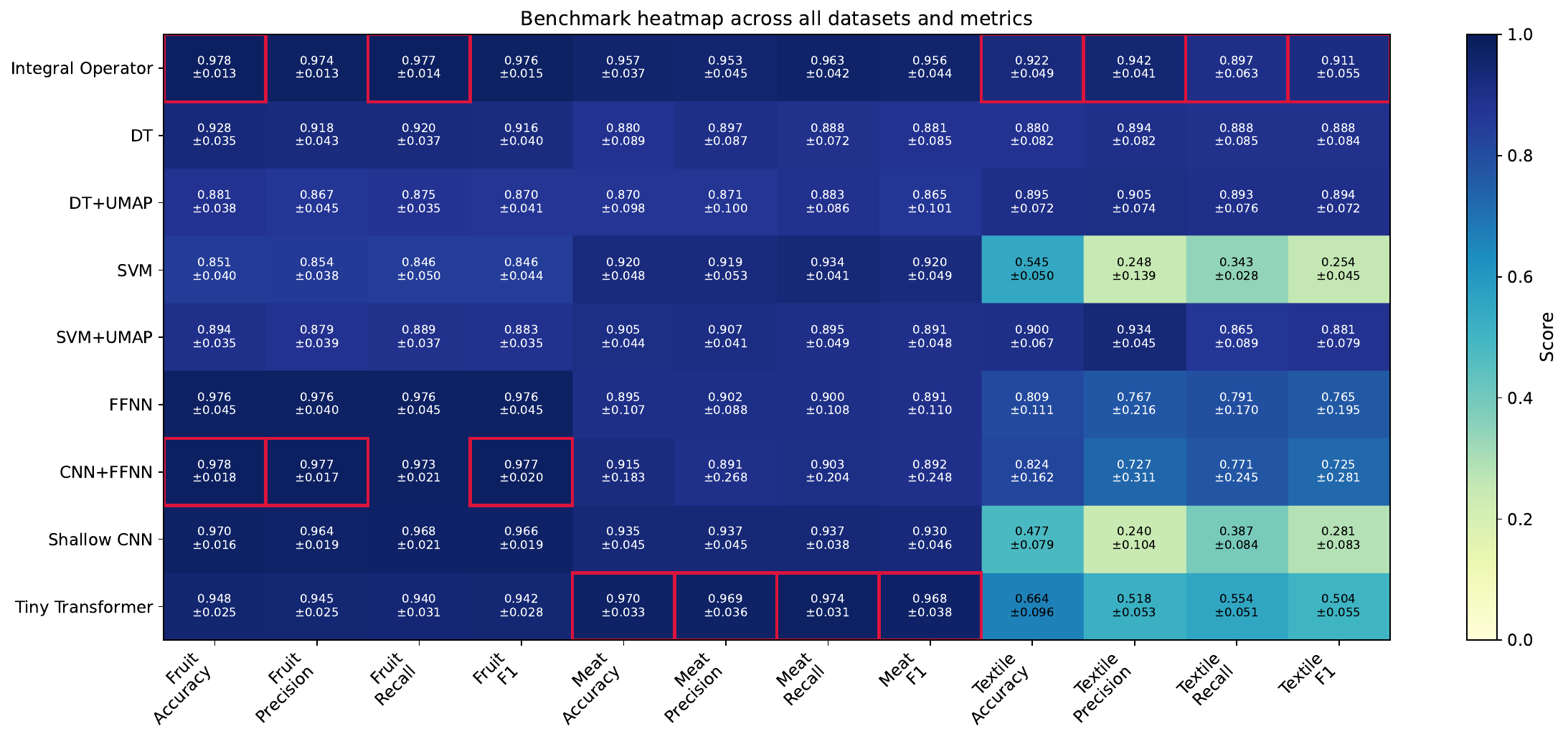}
    \caption{\textbf{Heatmap of all metrics across all datasets and models.} Each cell reports the mean $\pm$ standard deviation of the corresponding metric over the 10 Monte Carlo cross-validation splits. The same numerical values are reported in Table~\ref{tab:combined_benchmark_bold}.}
    \label{fig:benchmark_heatmap_all_metrics}
\end{figure}

\begin{figure}[t]
    \centering
    \includegraphics[width=\linewidth]{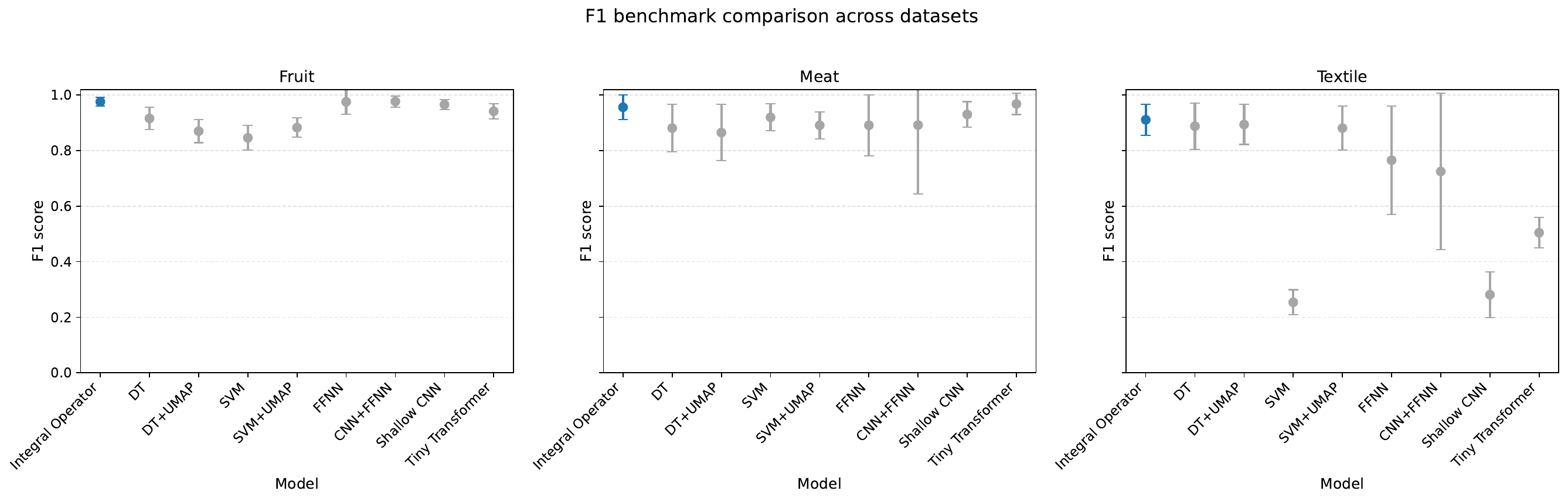}
    \caption{\textbf{Macro $F_{1}$ score across all models, dataset by dataset.} Error bars show $\pm 1$ standard deviation over the 10 Monte Carlo cross-validation splits.}
    \label{fig:f1_benchmark_comparison}
\end{figure}

Table~\ref{tab:combined_benchmark_bold} reports the full benchmark across the three datasets and four metrics, with means and standard deviations over the 10 Monte Carlo cross-validation splits. Best values per metric and dataset are highlighted in bold. Figure~\ref{fig:benchmark_heatmap_all_metrics} provides a heatmap of all 108 cells of the benchmark for visual inspection; Figure~\ref{fig:f1_benchmark_comparison} reports macro-$F_{1}$ with error bars per model per dataset. All values are reported on a scale of $1$. Therefore, for example, an accuracy of $0.80$ should read as $80\%$ accuracy. 

\begin{table*}[!t]
\centering
\caption{\textbf{Comprehensive benchmark across the three datasets.}
Best result per dataset and metric is shown in bold. Results are reported as macro-averaged performance (mean $\pm$ standard deviation) over 10 Monte Carlo cross-validation splits.}
\label{tab:combined_benchmark_bold}

\resizebox{\textwidth}{!}{%
\begin{tabular}{l l c c c c}
\toprule
Dataset & Model & Accuracy & Precision & Recall & F1 \\
\midrule

\textbf{Fruit}
& Integral Operator & $\mathbf{0.9780 \pm 0.0132}$ & $0.9740 \pm 0.0126$ & $\mathbf{0.9770 \pm 0.0142}$ & $0.9760 \pm 0.0151$ \\
& DT                & $0.9280 \pm 0.0349$ & $0.9180 \pm 0.0429$ & $0.9200 \pm 0.0374$ & $0.9160 \pm 0.0403$ \\
& DT+UMAP           & $0.8810 \pm 0.0381$ & $0.8670 \pm 0.0450$ & $0.8750 \pm 0.0347$ & $0.8700 \pm 0.0414$ \\
& SVM               & $0.8510 \pm 0.0404$ & $0.8540 \pm 0.0384$ & $0.8460 \pm 0.0504$ & $0.8460 \pm 0.0443$ \\
& SVM+UMAP          & $0.8940 \pm 0.0353$ & $0.8790 \pm 0.0390$ & $0.8890 \pm 0.0373$ & $0.8830 \pm 0.0353$ \\
& FFNN              & $0.9755 \pm 0.0450$ & $0.9764 \pm 0.0400$ & $0.9755 \pm 0.0450$ & $0.9755 \pm 0.0450$ \\
& CNN+FFNN          & $\mathbf{0.9780 \pm 0.0175}$ & $\mathbf{0.9770 \pm 0.0170}$ & $0.9730 \pm 0.0206$ & $\mathbf{0.9770 \pm 0.0200}$ \\
& Shallow CNN       & $0.9698 \pm 0.0158$ & $0.9641 \pm 0.0187$ & $0.9684 \pm 0.0208$ & $0.9657 \pm 0.0186$ \\
& Tiny Transformer  & $0.9479 \pm 0.0247$ & $0.9453 \pm 0.0251$ & $0.9402 \pm 0.0308$ & $0.9416 \pm 0.0276$ \\

\addlinespace[4pt]

\textbf{Meat}
& Integral Operator & $0.9570 \pm 0.0371$ & $0.9530 \pm 0.0452$ & $0.9630 \pm 0.0419$ & $0.9560 \pm 0.0438$ \\
& DT                & $0.8800 \pm 0.0888$ & $0.8970 \pm 0.0872$ & $0.8880 \pm 0.0725$ & $0.8810 \pm 0.0850$ \\
& DT+UMAP           & $0.8700 \pm 0.0978$ & $0.8710 \pm 0.0997$ & $0.8830 \pm 0.0860$ & $0.8650 \pm 0.1010$ \\
& SVM               & $0.9200 \pm 0.0483$ & $0.9190 \pm 0.0530$ & $0.9340 \pm 0.0409$ & $0.9200 \pm 0.0490$ \\
& SVM+UMAP          & $0.9050 \pm 0.0438$ & $0.9070 \pm 0.0414$ & $0.8950 \pm 0.0490$ & $0.8910 \pm 0.0484$ \\
& FFNN              & $0.8950 \pm 0.1066$ & $0.9022 \pm 0.0883$ & $0.9005 \pm 0.1084$ & $0.8914 \pm 0.1099$ \\
& CNN+FFNN          & $0.9150 \pm 0.1827$ & $0.8910 \pm 0.2684$ & $0.9030 \pm 0.2038$ & $0.8920 \pm 0.2482$ \\
& Shallow CNN       & $0.9350 \pm 0.0450$ & $0.9369 \pm 0.0454$ & $0.9373 \pm 0.0385$ & $0.9302 \pm 0.0461$ \\
& Tiny Transformer  & $\mathbf{0.9700 \pm 0.0332}$ & $\mathbf{0.9691 \pm 0.0361}$ & $\mathbf{0.9735 \pm 0.0310}$ & $\mathbf{0.9680 \pm 0.0382}$ \\

\addlinespace[4pt]

\textbf{Textile}
& Integral Operator & $\mathbf{0.9220 \pm 0.0487}$ & $\mathbf{0.9420 \pm 0.0408}$ & $\mathbf{0.8970 \pm 0.0629}$ & $\mathbf{0.9110 \pm 0.0551}$ \\
& DT                & $0.8800 \pm 0.0823$ & $0.8940 \pm 0.0817$ & $0.8880 \pm 0.0851$ & $0.8880 \pm 0.0839$ \\
& DT+UMAP           & $0.8950 \pm 0.0725$ & $0.9050 \pm 0.0740$ & $0.8930 \pm 0.0760$ & $0.8940 \pm 0.0718$ \\
& SVM               & $0.5450 \pm 0.0497$ & $0.2480 \pm 0.1390$ & $0.3430 \pm 0.0275$ & $0.2540 \pm 0.0450$ \\
& SVM+UMAP          & $0.9000 \pm 0.0667$ & $0.9340 \pm 0.0448$ & $0.8650 \pm 0.0891$ & $0.8810 \pm 0.0794$ \\
& FFNN              & $0.8091 \pm 0.1109$ & $0.7673 \pm 0.2160$ & $0.7911 \pm 0.1704$ & $0.7655 \pm 0.1949$ \\
& CNN+FFNN          & $0.8240 \pm 0.1617$ & $0.7270 \pm 0.3112$ & $0.7710 \pm 0.2451$ & $0.7250 \pm 0.2810$ \\
& Shallow CNN       & $0.4773 \pm 0.0794$ & $0.2401 \pm 0.1037$ & $0.3874 \pm 0.0837$ & $0.2813 \pm 0.0825$ \\
& Tiny Transformer  & $0.6636 \pm 0.0958$ & $0.5179 \pm 0.0528$ & $0.5541 \pm 0.0513$ & $0.5045 \pm 0.0546$ \\

\bottomrule
\end{tabular}%
}
\end{table*}

\paragraph{\textbf{High-level summary}}
Three observations stand out from the benchmark. First, \emph{no single model dominates across all datasets}: the best performers are CNN+FFNN on Fruit Purees (tied with NIO on accuracy), Tiny Transformer on Meat, and NIO on Textile. Second, \emph{the proposed NIO is the only model that achieves top-2 performance on all three datasets and all four metrics}: best or tied-best on Fruit Purees and Textile, and within $\sim 1.5\%$ of the best on Meat. Third, the relative performance gap between deep-learning baselines and NIO grows as the dataset becomes smaller and more complex, with the most pronounced separation on Textile, where CNN+FFNN drops to $0.824 \pm 0.162$ accuracy ($0.725 \pm 0.281$ $F_{1}$) and Tiny Transformer to $0.664 \pm 0.096$ accuracy, while NIO maintains $0.922 \pm 0.049$ accuracy ($0.911 \pm 0.055$ $F_{1}$).

\subsection{Per-dataset analysis}

\paragraph{\textbf{Fruit Purees (large, binary)}}
On the largest dataset, the NIO matches CNN+FFNN on accuracy ($0.978$) and is competitive across all metrics with all deep-learning baselines (FFNN, Shallow CNN, Tiny Transformer all within $\sim 0.005$ on $F_{1}$). Classical methods (DT, SVM, with or without UMAP) are markedly inferior, with the best classical model (SVM+UMAP) at $0.894$ accuracy. This is consistent with the established observation that deep learning extracts more discriminative features from FT-IR spectra when sufficient labelled data are available~\cite{zhang2022review,MishraDLNIR}. We note that the standard deviations of the NIO ($0.013$ on accuracy) are among the lowest in the deep-model group, suggesting that the stochastic regularization analysed in Section~\ref{subsec:MC_reg} is already operative in this comfort-zone regime.

\paragraph{\textbf{Meat (small, balanced, with outliers)}}
On the Meat dataset, the Tiny Transformer attains the best metrics (accuracy $0.970 \pm 0.033$, $F_{1}$ $0.968 \pm 0.038$), followed by the NIO (accuracy $0.957 \pm 0.037$, $F_{1}$ $0.956 \pm 0.044$) and the Shallow CNN ($0.935 \pm 0.045$). The full CNN+FFNN and the FFNN show much larger variance (standard deviations of $0.18$ and $0.11$ on accuracy respectively), revealing severe overfitting on individual splits despite competitive average means. The classical baselines (DT, DT+UMAP, SVM, SVM+UMAP) cluster around $0.87$--$0.92$ accuracy with intermediate variance. The fact that the Tiny Transformer outperforms NIO on this dataset is notable and we discuss it below in connection with the small class-conditional spectral overlap visible in Figure~\ref{fig:representative_spectra} (middle row).

\paragraph{\textbf{Textile (small, imbalanced, high overlap)}}
On the most challenging dataset, the NIO clearly outperforms every baseline on all four metrics: accuracy $0.922 \pm 0.049$ vs.\ SVM+UMAP $0.900 \pm 0.067$ and DT+UMAP $0.895 \pm 0.072$ (the two best baselines); deep learning baselines collapse, with FFNN at $0.809 \pm 0.111$ accuracy, CNN+FFNN at $0.824 \pm 0.162$, and Shallow CNN at $0.477 \pm 0.079$ (essentially failing to learn). The Tiny Transformer also drops sharply, to $0.664 \pm 0.096$ accuracy, confirming that attention-based architectures suffer in the high-overlap, small-imbalanced regime when no large-scale pre-training is available. The NIO improvement over SVM+UMAP is $+2.2\%$ in accuracy and $+3.0\%$ in $F_{1}$, and this is the only configuration in our benchmark in which a deep-learning model surpasses the strongest classical baseline on this dataset.

\paragraph{\textbf{When is the NIO framework the best model?}}
A consistent pattern across the three datasets is that the NIO's advantage is most pronounced where (i) the dataset is small, (ii) the class-conditional spectral overlap is high, and (iii) the deep baselines exhibit large run-to-run variance. Conversely, on the Meat dataset, where the class-conditional means show a clear separation between chicken and the other two classes across the full spectrum (Figure~\ref{fig:representative_spectra}, middle row), the small-data regime is partially mitigated by the discriminability of the inputs themselves; in this regime, a Tiny Transformer with strong tokenization can exploit the dominant feature directly without needing the additional inductive bias provided by the operator structure. This is consistent with the regularization-as-bias interpretation of Section~\ref{subsec:MC_reg}: when the data already separate well, the bias introduced by Monte Carlo integration brings no additional advantage, and slight underfitting may even reduce performance by a few percentage points. 

We emphasize that the computed $p$-values indicate that the top-performing models for each dataset are, in some cases, statistically equivalent. More specifically, on the fruit dataset, NIO is statistically equivalent to CNN, CNN+FFNN, and FFNN, while significantly outperforming the remaining models. On the meat dataset, NIO is statistically equivalent to the transformer architecture and significantly outperforms the other tested models. On the textile dataset, NIO is statistically equivalent to SVM+UMAP, while again outperforming the remaining models with statistical significance.

These results show that, although other models achieve performance comparable to NIO on individual datasets, NIO is the only model that consistently ranks among the top-performing methods across all tested datasets with statistical significance. In particular, no other model appears among the top-performing models, in terms of $p$-value comparisons across the tested metrics, on more than one dataset.

\subsection{Spectral interpretation}

While the proposed NIO is, like all deep-learning architectures, a black-box predictor, the class-conditional spectral statistics of Figure~\ref{fig:representative_spectra} offer a chemically meaningful lens through which to interpret the relative performance of the models. On the Fruit Purees FT-IR spectra (top row), strawberry and non-strawberry samples differ primarily in the magnitude and shape of absorption bands in the spectral indices 100--170, plausibly attributable to differences in carbohydrate and aromatic-acid composition. The fact that all deep baselines converge to $\sim 0.97$--$0.98$ accuracy here suggests that these spectral differences are accessible to relatively shallow models, with the operator-based inductive bias of NIO not providing a decisive additional advantage. On the Meat NIR spectra (middle row), the dominant feature is the systematic intensity shift of the chicken class across essentially the whole spectral range, plausibly attributable to lipid-content differences; once again, this feature is captured by a variety of architectures, hence the small spread of top scores. On the Textile NIR spectra (bottom row), the three classes overlap heavily within the $\pm 1$ standard-deviation bands across the entire 2000--2400 nm window, and the class means are largely concentric. This is the regime in which classical chemometric models (DT, SVM) and capacity-matched deep models (CNN+FFNN, Tiny Transformer) struggle, and in which the operator-learning bias (capturing nonlocal spectral interactions through the integrated kernel) combined with stochastic regularization provides the largest improvement. We refer the reader to~\cite{ContrerasXAI2024} for a complementary perspective on the explainability of spectroscopic deep-learning models.

\subsection{Comparison with related architectures}

A natural question concerns the position of NIO relative to recent neural operator and spectroscopic deep-learning architectures not included in the present benchmark. We make three observations. First, with respect to DeepONet~\cite{Lu2021} and Fourier Neural Operators~\cite{LiFNO,KovachkiNO}, the present work shares the operator-learning perspective but specifically targets first-kind integral equations and exploits Monte Carlo integration as a regularizer, an angle not emphasized in the prior literature. A direct head-to-head comparison would require adapting these architectures to the discrete-label classification setting (e.g.\ via a readout head on a function-valued output), which is conceptually straightforward but introduces design choices that lie outside the scope of this paper. Second, with respect to specialized spectroscopic CNN architectures such as RamanNet~\cite{RamanNet} and hybrid CNN--Transformer designs~\cite{RSTransformer}, our Shallow CNN, CNN+FFNN and Tiny Transformer baselines provide a capacity-matched cross-section that illustrates the typical behaviour of these families on small datasets; the more elaborate architectures of the cited works potentially attain higher accuracies on larger benchmark datasets, at the cost of greater parameter count and pre-training requirements. Third, with respect to small-sample strategies based on self-supervised pre-training~\cite{ZhaoNIRSSL,AbdalmalakfNIRS}, the NIO and these approaches are orthogonal and could in principle be combined: pre-training the encoder $E_{\phi_E}$ in a self-supervised fashion on unlabelled spectra is a natural avenue, which we discuss in Section~\ref{sec:limitations}.

\section{Limitations and Future Work}\label{sec:limitations}

The present study has some limitations, which we discuss explicitly to articulate concrete directions for follow-up work.

\paragraph{\textbf{Informal theoretical analysis of Monte Carlo regularization}}
The argument linking Monte Carlo integration in~\eqref{eqn:MC} to stochastic regularization (Section~\ref{subsec:MC_reg}) is informal. A formal analysis would quantify the effective regularization induced by integrand noise, ideally by deriving an equivalent kernel \emph{à la} \cite{SrivastavaDropout} for dropout, and by establishing generalization bounds for the resulting estimator. The dependence of the regularization strength on the kernel architecture, on the smoothness of $G_{\theta_G}$ in $\omega$, and on the budget $M_{\mathrm{train}}$ would also be valuable for guiding hyperparameter selection in new application domains.

\paragraph{\textbf{Absence of ablation studies on the Monte Carlo sampling budget}}
Our experiments are conducted at a single training Monte Carlo sampling budget $M_{\mathrm{train}} = 2000$ and a single evaluation budget $M_{\mathrm{test}} = 5000$. Systematic ablations over both budgets, and a head-to-head comparison with deterministic quadrature schemes (e.g.\ trapezoidal, Gauss--Legendre, Romberg) at matched compute cost, would directly test the regularization interpretation. We have observed qualitative consistency with this interpretation in preliminary runs (Section~\ref{subsec:MC_reg}), but a quantitative ablation is left to future work.

\paragraph{\textbf{Restricted set of deep-learning baselines}}
Our deep-learning baselines|FFNN, CNN + FFNN, Shallow CNN, Tiny Transformer|provide a capacity-matched cross-section of the most common architectural families in spectroscopic deep learning, but they do not exhaust the modern state of the art. In particular, we have not compared against full RamanNet implementations~\cite{RamanNet}, hybrid CNN--Transformer designs of the size used in Raman benchmarks~\cite{RSTransformer}, classification-adapted DeepONet or Fourier Neural Operators~\cite{Lu2021,LiFNO}, or self-supervised pre-training approaches~\cite{ZhaoNIRSSL,AbdalmalakfNIRS}. The latter, in combination with the NIO, is an especially natural extension: pre-training the encoder $E_{\phi_E}$ in a self-supervised fashion on unlabelled spectra would provide the operator-learning inductive bias of NIO with a stronger representational starting point.

\paragraph{\textbf{Independence assumption in cross-validation}}
The Monte Carlo cross-validation protocol of Section~\ref{subsec:protocol} assumes that the samples are exchangeable, which underlies the use of unstructured random 90/10 splits. If the underlying datasets contain instrumental replicates, batch effects or other forms of clustered structure, this assumption is violated and the reported variance underestimates the true generalization variability. Cluster-aware cross-validation (e.g.\ leave-one-batch-out) would address this issue, but requires knowledge of the data-acquisition structure that is not currently documented for all three datasets.

\paragraph{\textbf{Restriction to classification}}
The proposed framework is, in principle, equally applicable to regression: replacing the cross-entropy by mean squared error and the softmax by the identity yields a regression formulation in which the data function $f$ is a continuous target signal. Many inverse problems in spectroscopy (concentration prediction, quantitative imaging) are intrinsically regression problems, where PLS regression~\cite{Wold2001} and deep regression~\cite{QuantNIRDL} are the natural reference methods. We restrict the present empirical study to classification for clarity, and leave a regression-focused evaluation, including direct comparison with PLS regression on the same datasets where applicable, to future work.

\paragraph{\textbf{Cross-instrument and cross-domain generalization}}
We do not test cross-instrument transfer, which is a critical performance dimension in deployed spectroscopic systems. Standard transfer learning approaches~\cite{michelucci2024deep,kalatzis2023advanced} could plausibly be combined with the NIO by fine-tuning the encoder on a target instrument while keeping the operator kernel frozen, but this requires datasets with paired multi-instrument measurements that are not currently available to us.

\paragraph{\textbf{Limited domain scope}}
Although the proposed NIO is domain-agnostic and grounded in a general first-kind integral equation formulation~\cite{groetsch2007integral}, the empirical validation in this paper is restricted to spectroscopic classification. Applications to other inverse problems in soft computing, such as system identification, medical image reconstruction or geophysical inversion, are natural extensions, but they may surface architectural choices that go beyond the scope of the present work (e.g.\ multi-dimensional domains, time-dependent kernels). Investigating these extensions is, in our view, the most promising direction for future research on the proposed framework.

\section{Conclusions}\label{sec:conclusions}

We have introduced a Neural Integral Operator (NIO) framework for inverse problems formulated as integral equations of the first kind, in which the Urysohn kernel $G_{\theta_G}$ is parameterized by a feed-forward network, the latent function $u_{\phi_E}$ is produced by a convolutional encoder, and the resulting operator is evaluated by Monte Carlo integration. Three contributions stand out.

\paragraph{\textbf{(C1) An operator-learning architecture for first-kind integral equations}}
From a soft-computing perspective, the framework provides an explicit inductive bias on the structure of the input--output mapping by anchoring it to a first-kind integral equation. The kernel network and the encoder network are trained jointly end-to-end with a cross-entropy loss derived from the inverse-problem formulation, and the readout point $\sigma_E$ at which the operator is evaluated is itself learned from data. Compared with the existing neural operator literature, which has focused predominantly on forward problems and on second-kind equations, the proposed approach exposes the role of first-kind equations as a natural template for inverse classification and regression.

\paragraph{\textbf{(C2) Monte Carlo integration as implicit stochastic regularization}}
We argue, and support empirically, that Monte Carlo integration of the learned operator acts as an implicit stochastic regularizer, complementing parameter-level regularizers such as dropout and weight decay by injecting noise at the level of the integrand. This is, to our knowledge, a novel observation in the context of neural operator learning, and provides a principled explanation for the observed robustness of the NIO in low-data regimes. The asymmetric training/evaluation budget $M_{\mathrm{train}} = 2000 < M_{\mathrm{test}} = 5000$ used in our experiments is a direct corollary of this interpretation.

\paragraph{\textbf{(C3) Consistent top-2 performance across heterogeneous benchmarks}}
On three real-world spectroscopic classification benchmarks (FT-IR Fruit Purees, NIR Meat, NIR Textile) of varying size and complexity, the NIO is consistently among the top two performing models across all datasets and all four metrics considered (accuracy, macro precision, macro recall, macro $F_{1}$). It is the tied best performer also on the most challenging small-and-complex dataset (Textile). It is the only model that always appears among the top-performing models, in terms of $p$-value comparisons across the tested metrics, and no other model appears (in terms of statistical significance) on more than a single dataset. 

The empirical evidence supports the framing of the proposed approach as a viable soft-computing strategy for inverse problems in regimes where conventional deep learning approaches are limited by data scarcity. The methodology is domain-agnostic and applies wherever an inverse problem can be cast as a first-kind integral equation; we have validated it in the spectroscopic setting, but the underlying machinery is in no way specific to it. Promising directions for future work, articulated in Section~\ref{sec:limitations}, include a formal theoretical analysis of Monte Carlo regularization, regression extensions with PLS as a reference baseline, integration with self-supervised pre-training of the encoder, and applications to inverse problems in other domains.

\section*{Acknowledgments}
The authors would like to thank Dr. John Kalivas for inspiring conversations and for sharing the datasets used in this study. EZ acknowledges support from the NIH under grant R16GM154734. 

\bibliographystyle{alpha}
\bibliography{refs}

\begin{bibdiv}
\begin{biblist}

\bib{becht2019dimensionality}{article}{
      author={Becht, E.},
      author={McInnes, L.},
      author={Healy, J.},
      author={Dutertre, C.~A.},
      author={Kwok, I.~W.},
      author={Ng, L.~G.},
      author={Ginhoux, F.},
      author={Newell, E.~W.},
       title={Dimensionality reduction for visualizing single-cell data using umap},
        date={2019},
     journal={Nature Biotechnology},
      volume={37},
      number={1},
       pages={38\ndash 44},
}

\bib{blanco2022strategies}{article}{
      author={Blanco-Portals, Javier},
      author={Peir{\'o}, Francesca},
      author={Estrad{\'e}, S{\`o}nia},
       title={Strategies for eels data analysis. introducing umap and hdbscan for dimensionality reduction and clustering},
        date={2022},
     journal={Microscopy and Microanalysis},
      volume={28},
      number={1},
       pages={109\ndash 122},
}

\bib{ContrerasXAI2024}{article}{
      author={Contreras, Jhonatan},
      author={Bocklitz, Thomas},
       title={Explainable artificial intelligence for spectroscopy data: a review},
        date={2024},
     journal={Pfl{\"u}gers Archiv -- European Journal of Physiology},
      volume={477},
       pages={603\ndash 615},
}

\bib{del2023comparative}{article}{
      author={del Pozo-Bueno, Daniel},
      author={Kepaptsoglou, Demie},
      author={Peir{\'o}, Francesca},
      author={Estrad{\'e}, S{\`o}nia},
       title={Comparative of machine learning classification strategies for electron energy loss spectroscopy: Support vector machines and artificial neural networks},
        date={2023},
     journal={Ultramicroscopy},
      volume={253},
       pages={113828},
}

\bib{AbdalmalakfNIRS}{article}{
      author={Eastmond, Condell},
      author={Subedi, Aseem},
      author={De, Suvranu},
      author={Intes, Xavier},
       title={Deep learning in {fNIRS}: a review},
        date={2022},
     journal={Neurophotonics},
      volume={9},
      number={4},
       pages={041411},
}

\bib{gomez2021torchquad}{article}{
      author={G{\'o}mez, Pablo},
      author={Toftevaag, H{\aa}vard~Hem},
      author={Meoni, Gabriele},
       title={torchquad: Numerical integration in arbitrary dimensions with pytorch},
        date={2021},
     journal={Journal of Open Source Software},
      volume={6},
      number={64},
       pages={3439},
}

\bib{groetsch2007integral}{inproceedings}{
      author={Groetsch, Charles~W},
       title={Integral equations of the first kind, inverse problems and regularization: a crash course},
organization={IOP Publishing},
        date={2007},
   booktitle={Journal of physics: Conference series},
      volume={73},
       pages={012001},
}

\bib{holland1998use}{article}{
      author={Holland, James~K},
      author={Kemsley, E~Katherine},
      author={Wilson, Reginald~H},
       title={Use of fourier transform infrared spectroscopy and partial least squares regression for the detection of adulteration of strawberry purees},
        date={1998},
     journal={Journal of the Science of Food and Agriculture},
      volume={76},
      number={2},
       pages={263\ndash 269},
}

\bib{houston2020robust}{article}{
      author={Houston, J.},
      author={Glavin, F.G.},
      author={Madden, M.G.},
       title={Robust classification of high-dimensional spectroscopy data using deep learning and data synthesis},
        date={2020Mar},
     journal={Journal of Chemical Information and Modeling},
      volume={60},
      number={4},
       pages={1936\ndash 1954},
}

\bib{RamanNet}{article}{
      author={Ibtehaz, Nabil},
      author={Chowdhury, Muhammad E.~H.},
      author={Khandakar, Amith},
      author={Kiranyaz, Serkan},
      author={Rahman, M.~Sohel},
      author={Tahir, Anas},
      author={Qiblawey, Yazan},
      author={Rahman, Tawsifur},
       title={{RamanNet}: a generalized neural network architecture for {Raman} spectrum analysis},
        date={2023},
     journal={Neural Computing and Applications},
      volume={35},
      number={25},
       pages={18719\ndash 18735},
}

\bib{kalatzis2023advanced}{article}{
      author={Kalatzis, Dimitris},
      author={Spyratou, Ellas},
      author={Karnachoriti, Maria},
      author={Kouri, Maria~Anthi},
      author={Orfanoudakis, Spyros},
       title={Advanced raman spectroscopy based on transfer learning by using a convolutional neural network for personalized colorectal cancer diagnosis},
        date={2023},
     journal={Optics},
      volume={4},
      number={2},
       pages={310\ndash 320},
}

\bib{KovachkiNO}{article}{
      author={Kovachki, Nikola},
      author={Li, Zongyi},
      author={Liu, Burigede},
      author={Azizzadenesheli, Kamyar},
      author={Bhattacharya, Kaushik},
      author={Stuart, Andrew},
      author={Anandkumar, Anima},
       title={Neural operator: Learning maps between function spaces with applications to {PDE}s},
        date={2023},
     journal={Journal of Machine Learning Research},
      volume={24},
      number={89},
       pages={1\ndash 97},
}

\bib{LiFNO}{inproceedings}{
      author={Li, Zongyi},
      author={Kovachki, Nikola~B.},
      author={Azizzadenesheli, Kamyar},
      author={Liu, Burigede},
      author={Bhattacharya, Kaushik},
      author={Stuart, Andrew},
      author={Anandkumar, Anima},
       title={Fourier neural operator for parametric partial differential equations},
        date={2021},
   booktitle={International conference on learning representations (iclr)},
         url={https://openreview.net/forum?id=c8P9NQVtmnO},
}

\bib{Lu2021}{article}{
      author={Lu, Lu},
      author={Jin, Pengzhan},
      author={Pang, Guofei},
      author={Zhang, Zhiqiang},
      author={Karniadakis, George~Em},
       title={Learning nonlinear operators via deeponet based on the universal approximation theorem of operators},
        date={2021},
     journal={Nature Machine Intelligence},
      volume={3},
      number={3},
       pages={218\ndash 229},
}

\bib{luo2022deep}{article}{
      author={Luo, R.},
      author={Popp, J.},
      author={Bocklitz, T.},
       title={Deep learning for raman spectroscopy: A review},
        date={2022},
     journal={Analytica},
      volume={3},
      number={3},
       pages={287\ndash 301},
}

\bib{mark2010comparison}{article}{
      author={Mark, Howard},
      author={Rubinovitz, Ronald},
      author={Heaps, David},
      author={Gemperline, Paul},
      author={Dahm, Donald},
      author={Dahm, Kevin},
       title={Comparison of the use of volume fractions with other measures of concentration for quantitative spectroscopic calibration using the classical least squares method},
        date={2010},
     journal={Applied spectroscopy},
      volume={64},
      number={9},
       pages={995\ndash 1006},
}

\bib{mcinnes2018umap}{article}{
      author={McInnes, L.},
      author={Healy, J.},
      author={Melville, J.},
       title={Umap: Uniform manifold approximation and projection for dimension reduction},
        date={2018},
     journal={arXiv preprint arXiv:1802.03426},
}

\bib{michelucci2024deep}{article}{
      author={Michelucci, Umberto},
      author={Venturini, Francesca},
       title={Deep learning domain adaptation to understand physico-chemical processes from fluorescence spectroscopy small datasets and application to the oxidation of olive oil},
        date={2024},
     journal={Scientific Reports},
      volume={14},
      number={1},
       pages={22291},
}

\bib{QuantNIRDL}{article}{
      author={Mishra, Puneet},
      author={Passos, Dario},
       title={Realizing transfer learning for updating deep learning models of spectral data to be used in a new scenario},
        date={2021},
     journal={Chemometrics and Intelligent Laboratory Systems},
      volume={212},
       pages={104283},
}

\bib{MishraDLNIR}{article}{
      author={Mishra, Puneet},
      author={Passos, Dario},
      author={Marini, Federico},
      author={Xu, Junli},
      author={Amigo, Jos\'e~Manuel},
      author={Gowen, Aoife~A.},
      author={Jansen, Jeroen~J.},
      author={Biancolillo, Alessandra},
      author={Roger, Jean~Michel},
      author={Rutledge, Douglas~N.},
      author={Nordon, Alison},
       title={Deep learning for near-infrared spectral data modelling: Hypes and benefits},
        date={2022},
     journal={TrAC Trends in Analytical Chemistry},
      volume={157},
       pages={116804},
}

\bib{naes1989leverage}{article}{
      author={Næs, T.},
       title={Leverage and influence measures for principal component regression},
        date={1989},
     journal={Chemometrics and Intelligent Laboratory Systems},
      volume={5},
      number={2},
       pages={155\ndash 168},
}

\bib{SrivastavaDropout}{inproceedings}{
      author={Srivastava, Nitish},
      author={Hinton, Geoffrey},
      author={Krizhevsky, Alex},
      author={Sutskever, Ilya},
      author={Salakhutdinov, Ruslan},
       title={Dropout: A simple way to prevent neural networks from overfitting},
        date={2014},
      volume={15},
       pages={1929\ndash 1958},
}

\bib{Wold2001}{article}{
      author={Wold, Svante},
      author={Sj\"ostr\"om, Michael},
      author={Eriksson, Lennart},
       title={Pls-regression: a basic tool of chemometrics},
        date={2001},
     journal={Chemometrics and Intelligent Laboratory Systems},
      volume={58},
      number={2},
       pages={109\ndash 130},
}

\bib{NIDE}{inproceedings}{
      author={Zappala, Emanuele},
      author={Fonseca, Antonio H de~O},
      author={Moberly, Andrew~H},
      author={Higley, Michael~J},
      author={Abdallah, Chadi},
      author={Cardin, Jessica~A},
      author={van Dijk, David},
       title={Neural integro-differential equations},
        date={2023},
   booktitle={Proceedings of the aaai conference on artificial intelligence},
      volume={37},
       pages={11104\ndash 11112},
}

\bib{ANIE}{article}{
      author={Zappala, Emanuele},
      author={Fonseca, Antonio Henrique de~Oliveira},
      author={Caro, Josue~Ortega},
      author={Moberly, Andrew~Henry},
      author={Higley, Michael~James},
      author={Cardin, Jessica},
      author={Dijk, David~van},
       title={Learning integral operators via neural integral equations},
        date={2024},
     journal={Nature Machine Intelligence},
       pages={1\ndash 17},
}

\bib{RSTransformer}{article}{
      author={Zhang, Pengjie},
      author={Ren, Liangrui},
      author={Yang, Bingxu},
      author={Wang, Wuliji},
      author={Zhang, Xueying},
      author={Liu, Jia},
       title={A deep learning-based qualitative analysis model for spectroscopic signals combining convolutional neural network and transformer architecture},
        date={2023},
     journal={Measurement},
      volume={220},
       pages={113517},
}

\bib{zhang2022review}{article}{
      author={Zhang, Wenwen},
      author={Kasun, Liyanaarachchi~Chamara},
      author={Wang, Qi~Jie},
      author={Zheng, Yuanjin},
      author={Lin, Zhiping},
       title={A review of machine learning for near-infrared spectroscopy},
        date={2022},
     journal={Sensors},
      volume={22},
      number={24},
       pages={9764},
}

\bib{ZhaoNIRSSL}{article}{
      author={Zhao, Yong},
      author={Tian, Shilai},
      author={Yu, Long},
      author={Zhang, Zhongyou},
      author={Zhang, Weibing},
       title={A self-supervised contrastive learning framework with small samples for near-infrared spectroscopy},
        date={2024},
     journal={Analytical Methods},
      volume={16},
       pages={7059\ndash 7071},
}

\end{biblist}
\end{bibdiv}

\end{document}